\documentclass[12pt]{article}
\usepackage{tabularx}
\usepackage{amsmath,amssymb}
\usepackage{longtable}
\usepackage{graphicx,booktabs}
\usepackage[margin=1in]{geometry}
\usepackage{natbib}
\usepackage{subcaption} % in preamble
\usepackage{hyperref}
\usepackage{float,placeins}
\usepackage{threeparttable}
\usepackage{tikz}
\usepackage{siunitx}
\usepackage{subcaption}
\usetikzlibrary{positioning,arrows,shapes,calc}
\usepackage{pgfplots}
\pgfplotsset{compat=1.18}
\usepackage[activate=alltext]{microtype}

\title{Sleep-Based Homeostatic Regularization for Stabilizing Spike-Timing-Dependent Plasticity in Recurrent Spiking Neural Networks}
\author{Andreas Massey \and Aliaksandr Hubin \and Stefano Nichele \and Solve Sæbø}
\date{\today}

\begin{document}

\maketitle

\begin{abstract}
Spike-timing-dependent plasticity (STDP) provides a biologically-plausible learning mechanism for spiking neural networks (SNNs); however, Hebbian weight updates in architectures with recurrent connections suffer from pathological weight dynamics: unbounded growth, catastrophic forgetting, and loss of representational diversity. We propose a neuromorphic regularization scheme inspired by the synaptic homeostasis hypothesis: periodic offline phases during which external inputs are suppressed, synaptic weights undergo stochastic decay toward a homeostatic baseline, and spontaneous activity enables memory consolidation. We demonstrate that this sleep-wake cycle prevents weight saturation while preserving learned structure. Empirically, we find that low to intermediate sleep durations (10–20\% of training) improve stability on MNIST-like benchmarks in our STDP-SNN model, without any data-specific hyperparameter tuning. In contrast, the same sleep intervention yields no measurable benefit for the surrogate-gradient spiking neural network (SG-SNN). Taken together, these results suggest that periodic, sleep-based renormalization may represent a fundamental mechanism for stabilizing local Hebbian learning in neuromorphic systems, while also indicating that special care is required when integrating such protocols with existing gradient-based optimization methods.

\end{abstract}

\section{Introduction}

Spiking neural networks (SNNs) encode information through the spatiotemporal distribution of discrete spike events, enabling efficient event-driven computation on neuromorphic devices \citep{su_deep_2023}. This temporal sparsity yields orders-of-magnitude reductions in power consumption compared to dense, continuous-valued architectures \citep{isik_accelerating_2024}. However, the learning landscape for SNNs remains both inefficient and poorly understood \citep{akil_balanced_2021}, particularly concerning the stability and convergence of biologically plausible local learning rules \citep{stratton_making_2022}.

Spike-timing-dependent plasticity (STDP) emerges naturally as the canonical local learning rule for SNNs \citep{song_competitive_2000}. Weight modifications depend solely on the millisecond-scale timing differences between pre- and postsynaptic spikes: when presynaptic activity temporally precedes postsynaptic firing, weights strengthen; when the order reverses, weights weaken. This causality-respecting update rule is not only biologically faithful, but also theoretically motivated as it implements credit assignment through temporal adjacency and emerges from principles of spike coincidence detection \citep{jayabal_experience_2024, chauhan_emergence_2018}. Yet, in recurrent architectures essential for temporal processing and internal state maintenance, STDP exhibits severe pathologies. Recurrent connectivity amplifies initial weight perturbations through positive-feedback loops, driving three intertwined failure modes: (i) \textbf{weight saturation}, where unrestricted Hebbian potentiation of co-active synapses drives weights toward unbounded values or clipping limits \citep{gilson_stdp_2010}. The former creates catastrophic instabilities, while the latter eliminates further learning capacity; (ii) \textbf{representational collapse}, where weight-coupling induces spurious correlations that erode discriminative structure \citep{akil_balanced_2021}; and (iii) \textbf{catastrophic forgetting}, where sequential task learning overwrites previously-learned representations \citep{tadros_sleep-like_2022}.

The field has pursued two main approaches to stabilize STDP, each with significant costs. First, \textit{homeostatic mechanisms} employ global normalization schemes: synaptic scaling re-normalizes weight distributions toward firing-rate targets, intrinsic plasticity adapts neuronal thresholds, and structural plasticity modifies connectivity \citep{carlson_biologically_2013}. While effective, these approaches sacrifice strict locality - requiring global firing-rate statistics - and raise concerns about biological implausibility - synaptic scaling is typically modeled as multiplicative global factors rather than local mechanisms \citep{qiao_automatic_2016}. Second, \textit{surrogate-gradient} (SG) methods circumvent spike non-differentiability by replacing discontinuities with smooth membrane proxies during backpropagation, enabling error-based weight updates \citep{huh_gradient_2017}. SG methods can thus be coupled with standard $\ell_1$ or $\ell_2$ regularization and achieve solid accuracy for SNNs ($>$97\% on MNIST), but incur substantial costs that contradict the motivation for SNNs: they require error backpropagation (sacrificing locality), demand significant memory for gradient storage, utilize loss functions, and eliminate the event-driven sparsity that is arguably one of the strongest benefits of neuromorphic computation \citep{li_directly_2024, wang_evolving_2023}.

We propose a third approach grounded in the synaptic homeostasis hypothesis (SHy), a neuroscientific theory positing that sleep performs activity-weighted synaptic renormalization to maintain stability and consolidate learning \citep{tononi_sleep_2014}. During biological sleep, high-activity synapses undergo more aggressive down-scaling than low-activity synapses, achieving both weight renormalization and selective pruning of task-irrelevant connections, as explored in prior sleep-inspired neuromorphic models like \cite{tadros_sleep-like_2022, kinoshita_unified_2025, nere_sleep-dependent_2013}. We transpose this principle into a neuromorphic algorithm: periodic sleep/wake cycles similar to \cite{thiele_wake-sleep_2017}: During the sleep phase, (1) external inputs are suppressed, (2) synaptic weights undergo controlled power-law decay toward a homeostatic baseline, and (3) spontaneous neural activity, driven solely by intrinsic membrane noise inspired by \cite{ma_exploiting_2023} and \cite{deperrois_learning_2022}, permits continued STDP application, enabling consolidation through implicit replay. The latter has demonstrated improved intra-class consistency and reduced risks of overfitting to data \citep{deperrois_learning_2022, hoel_overfitted_2021, tadros_sleep-like_2022}.Critically, this mechanism preserves full locality (requiring only pre- and post-synaptic statistics), adds negligible computational overhead, and remains compatible with neuromorphic hardware. Moreover, we highlight a key distinction in how sleep-based homeostasis interacts with different learning paradigms. Sleep phases are mechanistically compatible with Hebbian learning, as both operate on local activity statistics. In contrast, few hybrid STDP/backpropagation (BP) approaches have been evaluated in SNNs, and existing methods typically attempt to circumvent locality constraints by introducing global reward signals or probabilistic learning rules. This reflects a fundamental incompatibility: STDP, by construction, cannot directly support BP’s reliance on global error gradients. Our experimental results are consistent with this view. While spontaneous activity and noisy weight reductions during sleep may, in principle, aid surrogate-gradient-trained SNNs in escaping poor local optima, we find no evidence that this translates into systematic improvements in learning performance\footnote{There have been some attempts to combine surrogate-methods with STDP, for example, the noisy-SNN (NSNN) model developed by \cite{ma_exploiting_2023}. They leverage noisy membrane activity to ensure a more robust network and feed the loss into the membrane potential, similar to other surrogate-methods.}. 

Our contributions are threefold. First, we present a principled sleep-wake mechanism combining homeostatic weight decay with spontaneous replay, deriving stability conditions showing that periodic renormalization bounds weight growth and maintains representational structure, consistent with \citep{capone_sleep-like_2019}. Second, we suggest drivers to prevent performance-saturation: consolidation-through-replay, and maintained homeostatic balance—with mechanistic explanations. Third, we provide empirical validation on MNIST-family benchmarks, showing that intermediate sleep durations (10\%–20\% of baseline training) reliably improve performance in our STDP-based model, whereas no sleep regime yields gains for the surrogate-gradient SNN model. 

The remainder of the paper is organized as follows: Section~2 describes the methods used in this work, including the two spiking neural network implementations under study, their architectures, learning rules, and sleep–wake regularization protocols. As an initial validation step, we first evaluate the STDP-SNN on a controlled toy classification task to verify stable learning dynamics under unsupervised STDP, before scaling to MNIST-family benchmarks with both models. Section~3 presents the results. Section~4 discusses the broader implications for neuromorphic hardware and future research directions.

\section{Methods}
\subsection{Overview}
We compare two spiking neural network implementations that differ in architecture and learning paradigm, but are both augmented with the same sleep-inspired homeostatic regularization mechanism. The first model, the spike-timing-dependent plasticity spiking neural network (STDP-SNN), is a recurrent excitatory–inhibitory architecture trained using local STDP and iSTDP learning rules. It is implemented from scratch in \verb|numpy| with \verb|numba| acceleration and explicitly models leaky integrate-and-fire neuron dynamics, synaptic plasticity, and sleep–wake cycles. During sleep phases, synaptic weights undergo noise-driven renormalization while STDP remains active, enabling replay-like consolidation.

The second model, the surrogate-gradient spiking neural network (SG-SNN), is implemented using the \verb|snntorch| framework and consists of feedforward linear layers coupled with leaky integrate-and-fire units, trained via surrogate-gradient backpropagation. In this setting, sleep is introduced by periodically interrupting task-driven training to apply the same weight-decay and noise mechanisms, but without ongoing STDP. 

Consequently, sleep plays a distinct functional role in the two models: facilitating replay-based consolidation in the STDP-SNN, while acting solely as a stochastic regularizer in the surrogate-gradient regime.

All code for the STDP-SNN model is publicly available at \cite{massey_spikingneuralnetwork_2025}.
\subsection{STDP-model}
The STDP-SNN serves as our primary biologically inspired model, combining local spike-timing-dependent plasticity with a periodic, sleep-inspired homeostatic regularization mechanism. The network is trained in an unsupervised manner using exclusively local learning rules, while stability is enforced through noisy weight renormalization during interleaved sleep phases. This design allows the model to learn class-structured representations while avoiding the pathological weight dynamics typically associated with recurrent STDP-based networks. 
\subsubsection{Architecture}
To construct our network, we drew inspiration from prior work on unsupervised clustering in spiking neural networks \citep{zenke_diverse_2015}, which demonstrated that recurrent excitatory–inhibitory architectures trained with STDP can self-organize class-selective representations. This class of models is particularly well suited for studying stability in Hebbian learning, as recurrent excitation amplifies small weight perturbations and is therefore prone to pathological dynamics in the absence of explicit regularization. By adopting a similar architecture, we intentionally place our model in a challenging regime where uncontrolled STDP is known to fail, allowing us to directly assess whether sleep-inspired homeostatic mechanisms can stabilize learning without sacrificing locality.

To keep the results comparable, we chose the following network parameters: a Poisson process input encoding into spikes layer ($N_{\text{in}}=225$ neurons), a recurrent excitatory readout layer ($N_{\text{exc}}=200$), and a lateral inhibitory layer ($N_{\text{inh}}=50$). Note that \cite{zenke_diverse_2015} also included a recurrent weight within the lateral inhibitory layer. We opted out of this due to increased complexity in estimating hyper-parameters. Connectivity is random and sparse: $P_{\text{in}\to\text{exc}}=10\%$, $P_{\text{exc}\to\text{exc}}=15\%$, $P_{\text{exc}\to\text{inh}}=20\%$, $P_{\text{inh}\to\text{exc}}=25\%$, with no self-connections. This architecture captures key sensory cortex properties: feedforward thalamic input, recurrent excitatory dynamics, and local inhibitory control.
\begin{figure}[h]
\centering
\begin{tikzpicture}[scale=1.1, line width=2pt, font=\normalsize]

% Input box
\draw (0, 0) rectangle (3, 1.5) node[midway, align=center] {Input(Poisson)\\ $N = 225$};

% Excitatory box
\draw (5, 0) rectangle (8, 1.5) node[midway, align=center] {Excitatory\\ $N_P=200$};

% Inhibitory box  
\draw (10, 0) rectangle (13, 1.5) node[midway, align=center] {Inhibitory\\ $N = 50$};

% Forward arrows
\draw[->, line width=2pt] (3, 0.75) -- (5, 0.75) node[midway, above, yshift=0.15cm] {\small $P = 10\%$};
\draw[->, line width=2pt] (8, 0.75) -- (10, 0.75) node[midway, above, yshift=0.15cm] {\small $P = 20\%$};

% Recurrent self-connection (loop on top)
\draw[->, line width=2pt] (6, 1.5) .. controls (6, 2.3) and (7, 2.3) .. (7, 1.5) node[midway, above, yshift=0.2cm] {\small $P=15\%$};

% Inhibitory feedback (bottom loop)
\draw[->, line width=2pt] (11.5, 0) .. controls (11.5, -0.6) and (6.5, -0.6) .. (6.5, 0) node[midway, below, yshift=-0.3cm] {\small $P=25\%$};
\end{tikzpicture}
\caption{Network architecture: three-layer SNN with feedforward input ($P=10\%$), recurrent excitation ($P=15\%$), and lateral inhibition ($P=25\%$). Sparse connectivity enables event-driven computation.}
\label{fig:architecture}
\end{figure}
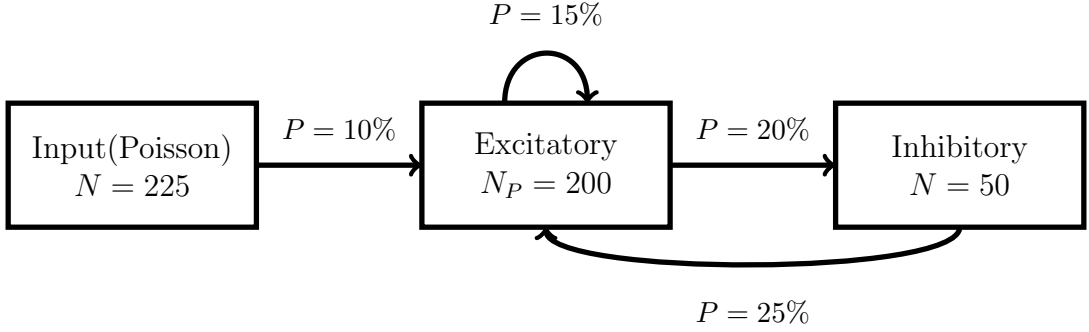

\subsubsection{Neuron model}
Each neuron integrates synaptic inputs via leaky integrate-and-fire (LIF) dynamics:
\begin{equation}
\tau_m \frac{dU_i(t)}{dt} = -(U_i(t) - U_{\text{rest}}) + R_m I_i^{\text{syn}}(t) + \xi_i(t),
\label{eq:lif}
\end{equation}
where $U_i(t)$ is the membrane potential, $\tau_m$ is the time constant, $R_m$ is the membrane resistance, $U_{\text{rest}}$ is the resting potential, $I_i^{\text{syn}}(t) = \sum_j w_{ij} S_j(t)$ is the synaptic current, and $\xi_i(t) \sim \mathcal{N}(0,\sigma^2)$ is intrinsic noise. When $U_i(t) \geq U_{\text{th}}^i(t)$, neuron $i$ emits a spike ($S_i(t)=1$) and resets to $U_{\text{reset}}$. The spike threshold is adaptive (preventing runaway firing):
\begin{align*}
&\frac{dU_{\text{th}}^i(t)}{dt} = U_{\text{th}}(0)+\alpha_i,
\label{eq:threshold}
\\
    &\frac{d\alpha_i}{dt} = -\frac{\alpha_i}{\tau_{th}}+S_i(t)\delta,
\end{align*}
with baseline spike threshold $U_{\text{th}}(0)$, time constant $\tau_{\text{th}}$, and jump amplitude $\delta$. For example, with $\delta = 3$ we reduce the probability of prolonged spike trains, i.e., each time neuron $i$ spikes, the spiking threshold increases by 3 mV ($\sim6.5\%$ increase per postsynaptic spike). 

\subsubsection{Learning rule}
Excitatory synapses update via spike-timing-dependent plasticity (STDP):
\begin{equation}
\Delta w_{ij}^{\text{exc}}(\Delta t) = \eta_{\text{exc}} \begin{cases}
A_+ \exp(-\Delta t/\tau_+) & \text{if } \Delta t \geq 0\\
-A_- \exp(\Delta t/\tau_-) & \text{otherwise},
\end{cases}
\label{eq:stdp}
\end{equation}
where $\Delta t = t_{\text{post}} - t_{\text{pre}}$ is spike timing, $\eta_{\text{exc}}$ is the learning rate, and $A_\pm$, $\tau_\pm$ are learning amplitudes and time constants. Inhibitory synapses employ inverted STDP (iSTDP):
\begin{equation}
\Delta w_{ij}^{\text{inh}}(\Delta t) = \eta_{\text{inh}} \begin{cases}
-A_+ \exp(-\Delta t/\tau_+) & \text{if } \Delta t \geq 0\\
A_- \exp(\Delta t/\tau_-) & \text{otherwise},
\end{cases}
\label{eq:istdp}
\end{equation}
with $\eta_{\text{inh}}$ being the (un-)learning rate. iSTDP ensures inhibitory neurons specialize for highly-active excitatory neurons, providing homeostatic load-balancing.

\begin{figure}[h]
\centering
\begin{tikzpicture}[scale=0.95]
\begin{axis}[
  axis lines=left,
  xlabel=$\Delta t$ (milliseconds),
  ylabel=$\Delta w$ (normalized),
  width=10cm,
  height=6cm,
  xmin=-100, xmax=100,
  ymin=-0.3, ymax=0.3,
  xtick={-100,-50,0,50,100},
  ytick={-0.25,0,0.25},
  grid=major,
  grid style={solid, gray, line width=0.4pt, opacity=0.5},
  xlabel style={font=\small},
  ylabel style={font=\small},
  tick label style={font=\small}
]

% Potentiation curve
\addplot[thick, solid, black, domain=0:100, samples=60] {0.25 * exp(-x/20)};

% Depression curve
\addplot[thick, solid, black, domain=-100:0, samples=60] {-0.2 * exp(x/20)};

% Zero line
\addplot[thin, dashed, black, domain=-100:100] {0};

\node[anchor=south west, font=\small] at (axis cs: 40, 0.15) {Potentiation};
\node[anchor=north east, font=\small] at (axis cs: -40, -0.15) {Depression};

\end{axis}
\end{tikzpicture}
\caption{STDP learning window: weights strengthen (positive $\Delta t$) when presynaptic spike precedes postsynaptic, weaken (negative $\Delta t$) otherwise.}
\label{fig:stdp_curve}
\end{figure}
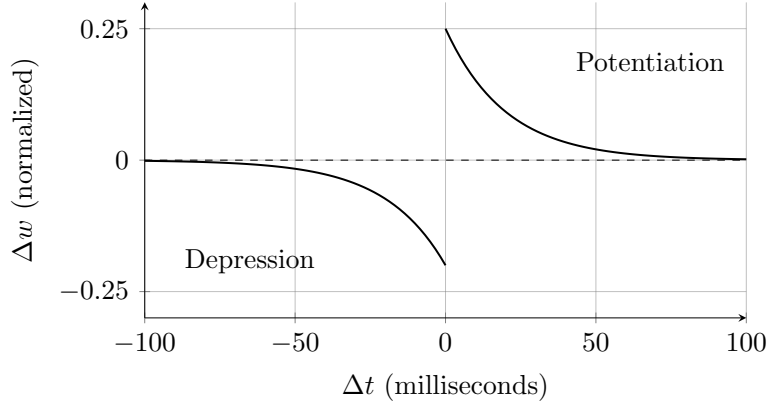

\subsubsection{Sleep-protocol}\label{sec:sp_SG}
The core innovation of this article is that we implemented a periodic sleep--wake schedule, controlled by the \textit{sleep-ratio}. Time is partitioned into intervals of set lengths, \textit{sleep interval}. At the start of each interval, the simulator enters a \emph{hard-pause} sleep phase and performs up to
\[
T_{\mathrm{sleep}}=\max\!\left(1,\left\lfloor \texttt{sleep-ratio}\cdot \texttt{sleep interval}\right\rceil\right)
\]
\emph{virtual} sleep iterations without advancing the real simulation time.

During sleep, weights decay via power-law dynamics that preserve relative ordering:
\begin{equation}
w_i(t+1) = w_{\text{tgt}} \left(\frac{w_i(t)}{w_{\text{tgt}}}\right)^\lambda,
\label{eq:decay}
\end{equation}
where $w_{\text{tgt}}$ is the target to which the function is converging and $\lambda$ denotes the decay rate. 
During sleep, sensory input is suppressed (all stimulus spikes are set to zero), and membrane dynamics are driven by intrinsic noise, producing spontaneous spiking activity that can continue to elicit STDP updates. In addition, synaptic weights are renormalized toward target magnitudes via the sleep operator. 

Sleep terminates when the sum of weights $\mathcal{W}_{x}(t)$ reaches $w^{\min}_x$

\begin{equation}
\mathcal{W}_x(t)  \leq w^{\min}_x = \alpha_{\text{base}} \cdot \mathcal{W}_{x}(0), \quad \alpha_{\text{base}}<\alpha_{\text{trig}}
\label{eq:wakeup}
\end{equation}

or when the \textit{sleep interval} is surpassed without reaching $w^{\min}_x$.

\begin{figure}[h!]
\centering
\begin{tikzpicture}[scale=1.0]
\begin{axis}[
  axis lines=left,
  xlabel=Time (simulation steps),
  ylabel={$\mathcal{W}(t)$ (total weight)},
  width=12cm,
  height=6cm,
  xmin=0, xmax=1000,
  ymin=30, ymax=145,
  xtick={0,200,400,600,800,1000},
  ytick={40,80,120},
  grid=major,
  grid style={solid, gray, line width=0.4pt, opacity=0.5},
  xlabel style={font=\small},
  ylabel style={font=\small},
  tick label style={font=\small}
]

% Wake phase 1: clean linear growth
\addplot[thick, solid, black, domain=0:200, samples=100] {50 + 0.37*x};

% Sleep phase 1: exponential decay 
\addplot[thick, solid, black, samples=80, domain=200:350, forget plot] 
  table[row sep=\\, y expr=130 * (0.98)^(x-200) + (rand-0.5)*7] {
x \\ 200 \\ 208 \\ 216 \\ 224 \\ 232 \\ 240 \\ 248 \\ 256 \\ 264 \\ 272 \\ 280 \\ 288 \\ 296 \\ 304 \\ 312 \\ 320 \\ 328 \\ 336 \\ 344 \\ 350 \\
};

% Wake phase 2
\addplot[thick, solid, black, domain=350:550, samples=100] {60 + 0.33*(x-350)};

% Sleep phase 2
\addplot[thick, solid, black, samples=80, domain=550:700, forget plot] 
  table[row sep=\\, y expr=130 * (0.98)^(x-550) + (rand-0.5)*7] {
x \\ 550 \\ 558 \\ 566 \\ 574 \\ 582 \\ 590 \\ 598 \\ 606 \\ 614 \\ 622 \\ 630 \\ 638 \\ 646 \\ 654 \\ 662 \\ 670 \\ 678 \\ 686 \\ 694 \\ 700 \\
};

% Wake phase 3
\addplot[thick, solid, black, domain=700:1000, samples=100] {65 + 0.3*(x-800)};

% Upper threshold (w_max)
\draw[thick, dashed, black] (0, 130) -- (1000, 130) node[right, xshift=-3.5cm, yshift=0.1cm, font=\small] {$w_{\max}$};

% Lower threshold (w_min)
\draw[thick, dashed, black] (0, 100) -- (1000, 100) node[right, xshift=-3.5cm, yshift=0.1cm, font=\small] {$w_{\min}$};

% Phase labels
\node[above, font=\small, fill=white, opacity=0.85] at (axis cs: 100, 142) {Wake};
\node[above, font=\small, fill=white, opacity=0.85] at (axis cs: 275, 105) {Sleep $+$ noise};
\node[above, font=\small, fill=white, opacity=0.85] at (axis cs: 625, 142) {Wake};
\node[above, font=\small, fill=white, opacity=0.85] at (axis cs: 625, 105) {Sleep $+$ noise};
\node[above, font=\small, fill=white, opacity=0.85] at (axis cs: 850, 142) {Wake};

\end{axis}
\end{tikzpicture}
\caption{
Weight dynamics: wake phases show deterministic linear STDP-driven growth; sleep phases show power-law regularization, Eq.~\ref{eq:decay} with superimposed random noise resembling realizations of Gaussian random variables, representing spontaneous neural activity during consolidation. System cycles robustly between wake (input-driven) and sleep (homeostatic noisy decay), maintaining weights within plausible intervals.
}
\label{fig:weight_dynamics}
\end{figure}
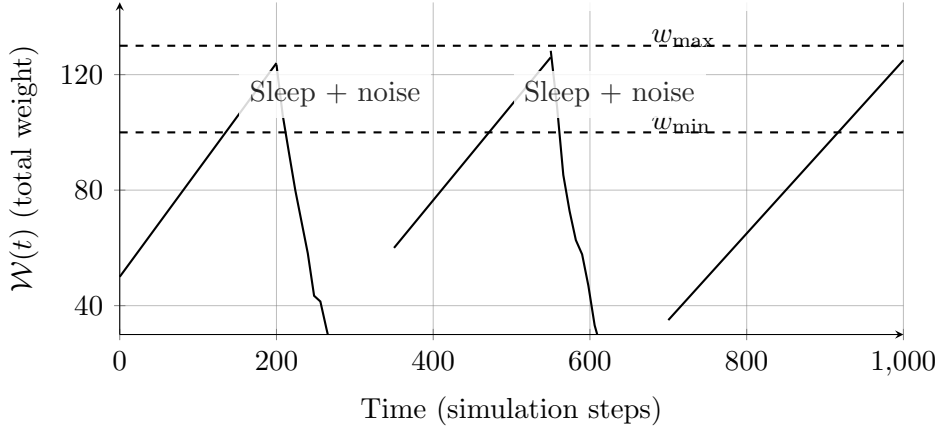

\subsubsection{Evaluation}
To compute the accuracy, we first aggregated the average spike rates for each training sample, and standardized the feature vector into $\mu=0$ and unit variance. Then, we perform dimensionality reduction on the standardized feature vector while maintaining 95\% variance with PCA. Finally, we train a multinomial logistic regression (MLR) on the transformed features. The trained classifiers can then predict classes on the unseen test-data processed by the network. 

\subsection{Toy geometric experiment}
To assess the clustering behavior of the unsupervised STDP-SNN and to isolate the effects of sleep-based regularization from dataset-specific confounds, we designed a controlled toy classification task. This experiment serves as a diagnostic validation, allowing us to verify that the network develops stable and discriminative internal representations under local Hebbian learning before evaluating performance on large-scale image benchmarks.

\subsubsection{Data}
\begin{figure}[h]
    \centering

    % --- Left figure ---
    \begin{subfigure}[b]{0.45\linewidth}
        \centering
        \includegraphics[width=0.7\linewidth]{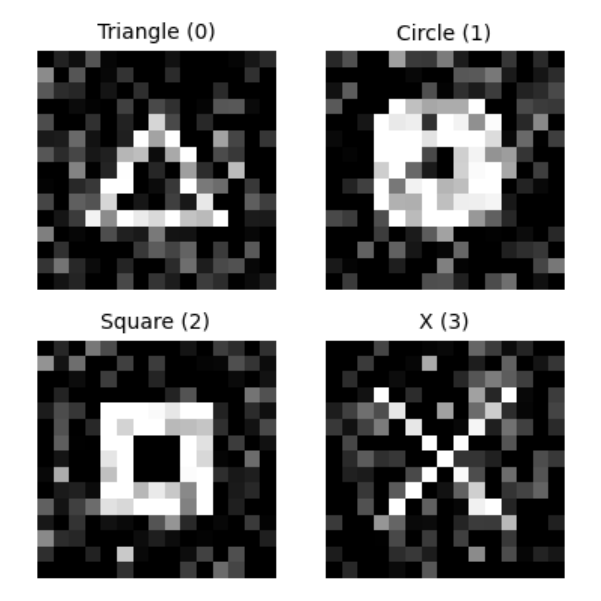}
        \caption{}   % unnumbered caption
        \label{fig:geomfig_data}
    \end{subfigure}
    % --- Right figure ---
    \begin{subfigure}[b]{0.45\linewidth}
        \centering
        \includegraphics[width=0.9\linewidth]{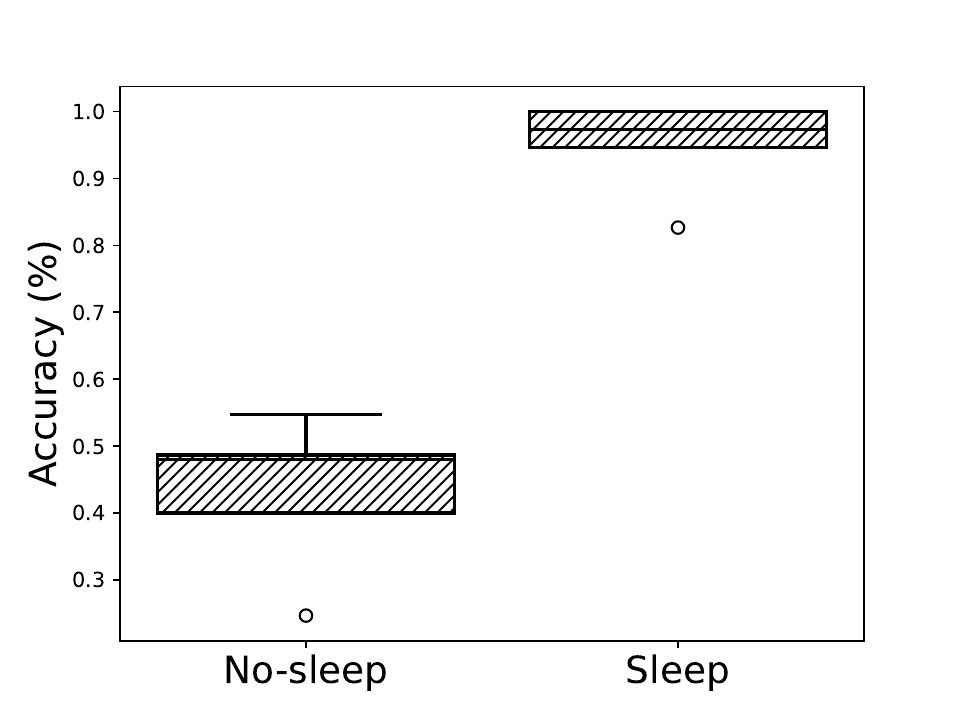}
        \caption{}   % unnumbered caption
        \label{fig:sleep-no-sleep-comp}
    \end{subfigure}

    \caption{(a) training set composed of four geometric classes (triangle, circle, square, and cross), each subjected to Gaussian noise $\mathcal{N}(0,0.02)$. All training examples use the same base shapes, with noise as the stochastic component introducing sample-by-sample variance. (b) shows accuracy attained with the \textit{no-sleep} and \textit{sleep} protocol.}
\end{figure}

We used a synthetic geometric dataset consisting of $7{,}100$ grayscale images at $15\times15$ resolution (Fig.~\ref{fig:geomfig_data}). Each image belongs to one of four geometric classes and is corrupted by additive Gaussian noise $\zeta_i \sim \mathcal{N}(0, 0.02)$ to introduce controlled variability while preserving class identity. Each stimulus was presented to the network for $100$ ms.

The dataset was split into $6{,}000/100/1{,}000$ samples for training, validation, and testing, respectively. Mini-batches were constructed to contain balanced class representations, ensuring that learning dynamics reflect clustering behavior rather than class-frequency effects.

\subsubsection{Design}
To estimate the efficacy of sleep, we compare model performance under two training regimes: sleep and no sleep. During the sleep condition, synaptic weights undergo periodic regularization, whereas in the no-sleep condition training proceeds continuously without these phases. In both regimes, synaptic weights are constrained to remain within their respective functional groups—excitatory or inhibitory—through sign-clamping, a mechanism that enforces fixed polarity by preventing weights from crossing the zero boundary and thereby changing their postsynaptic effect.

\subsubsection{Results}
\begin{figure}
    \centering
    % --- Left figure ---
    \begin{subfigure}[b]{0.45\linewidth}
        \centering
        \includegraphics[width=\linewidth]{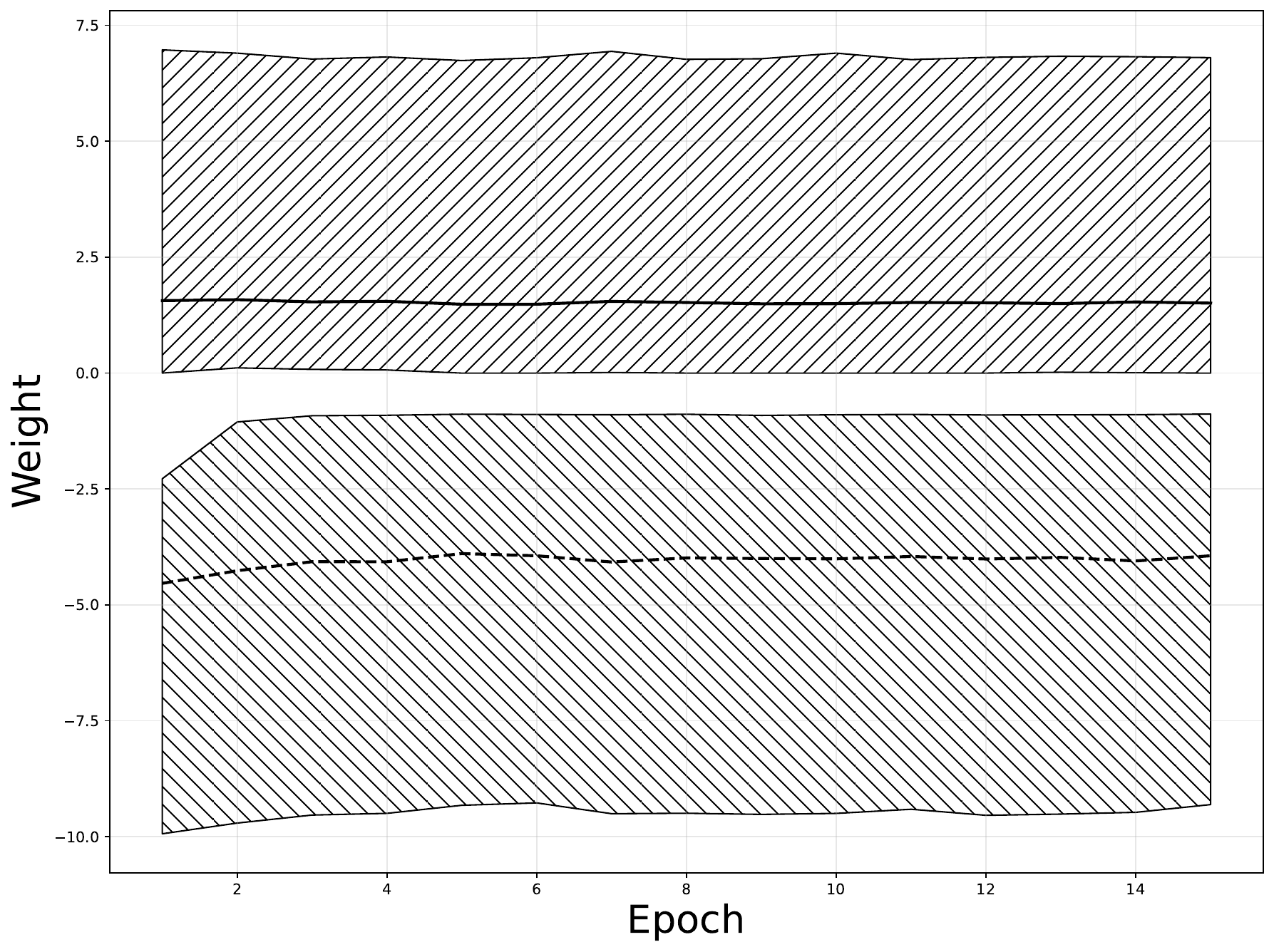}
        \caption{}   % unnumbered caption
        \label{fig:w_evol_sleep}
    \end{subfigure}
    \hfill
    % --- Right figure ---
    \begin{subfigure}[b]{0.45\linewidth}
        \centering
        \includegraphics[width=\linewidth]{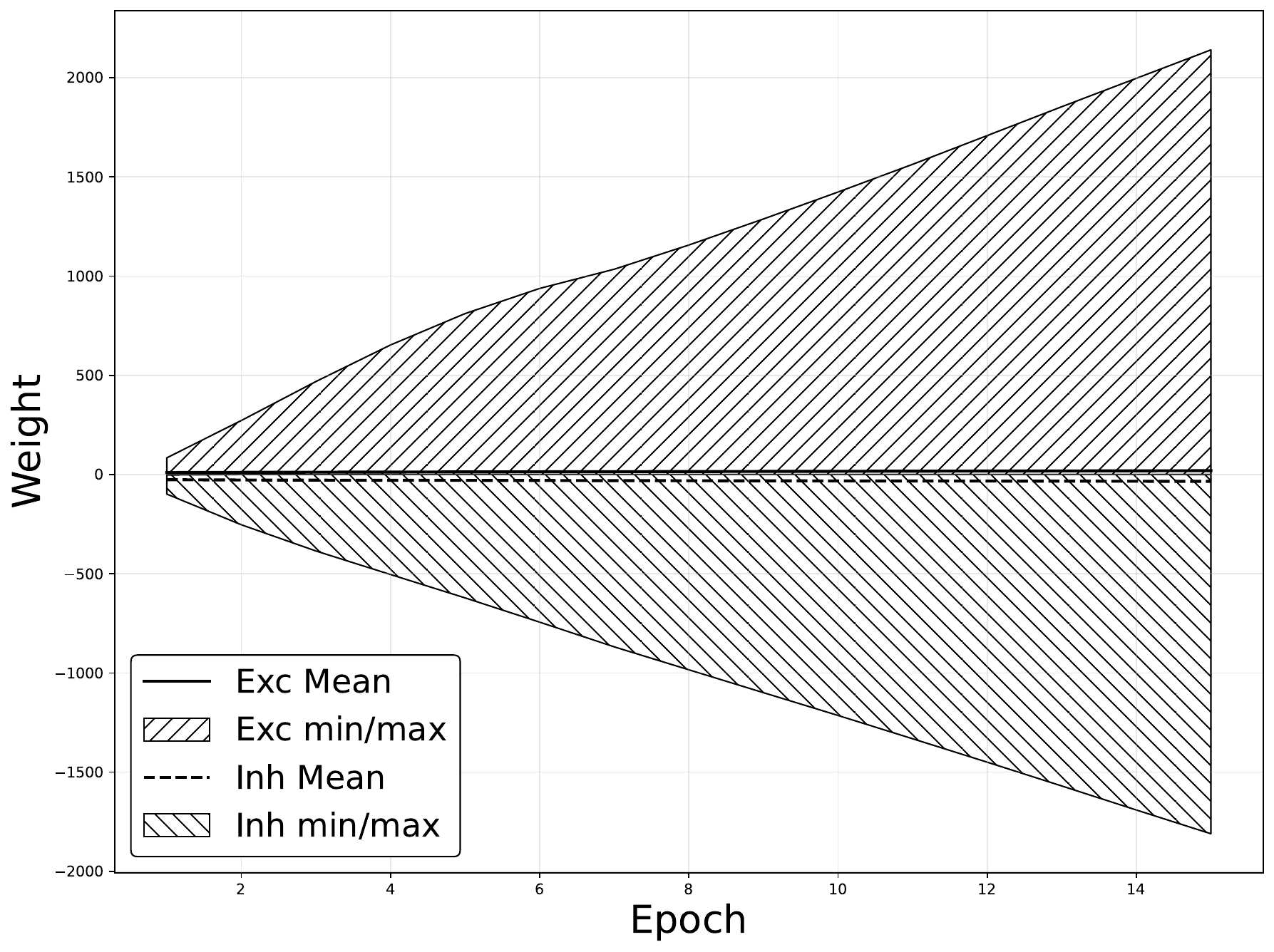}
        \caption{}   % unnumbered caption
        \label{fig:w_evol_no_sleep}
    \end{subfigure}

    % --- Left figure ---
    \begin{subfigure}[b]{0.45\linewidth}
        \centering
        \includegraphics[width=\linewidth]{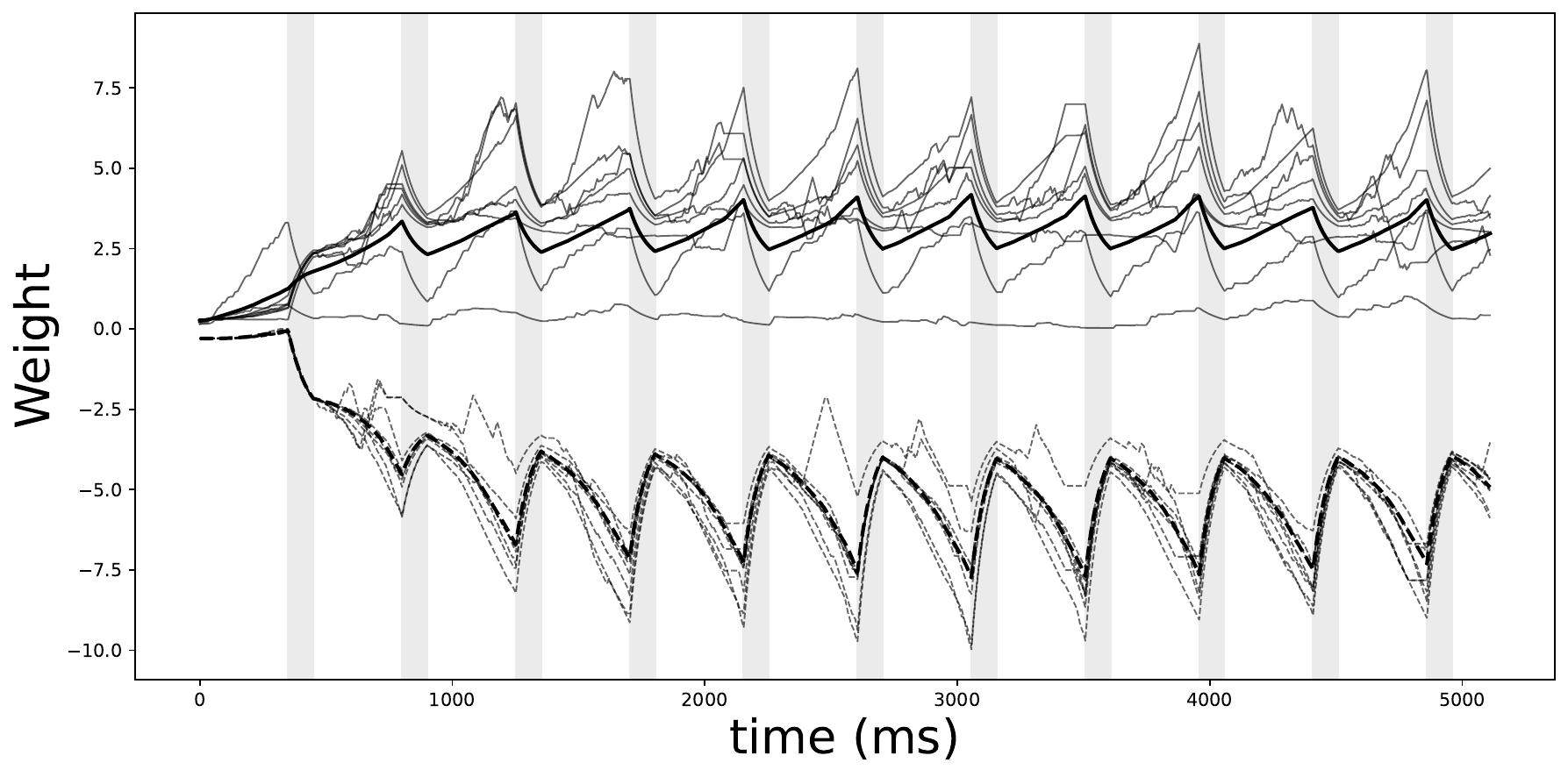}
        \caption{}   % unnumbered caption
        \label{fig:w_epo_sleep}
    \end{subfigure}
    \hfill
    % --- Right figure ---
    \begin{subfigure}[b]{0.45\linewidth}
        \centering
        \includegraphics[width=\linewidth]{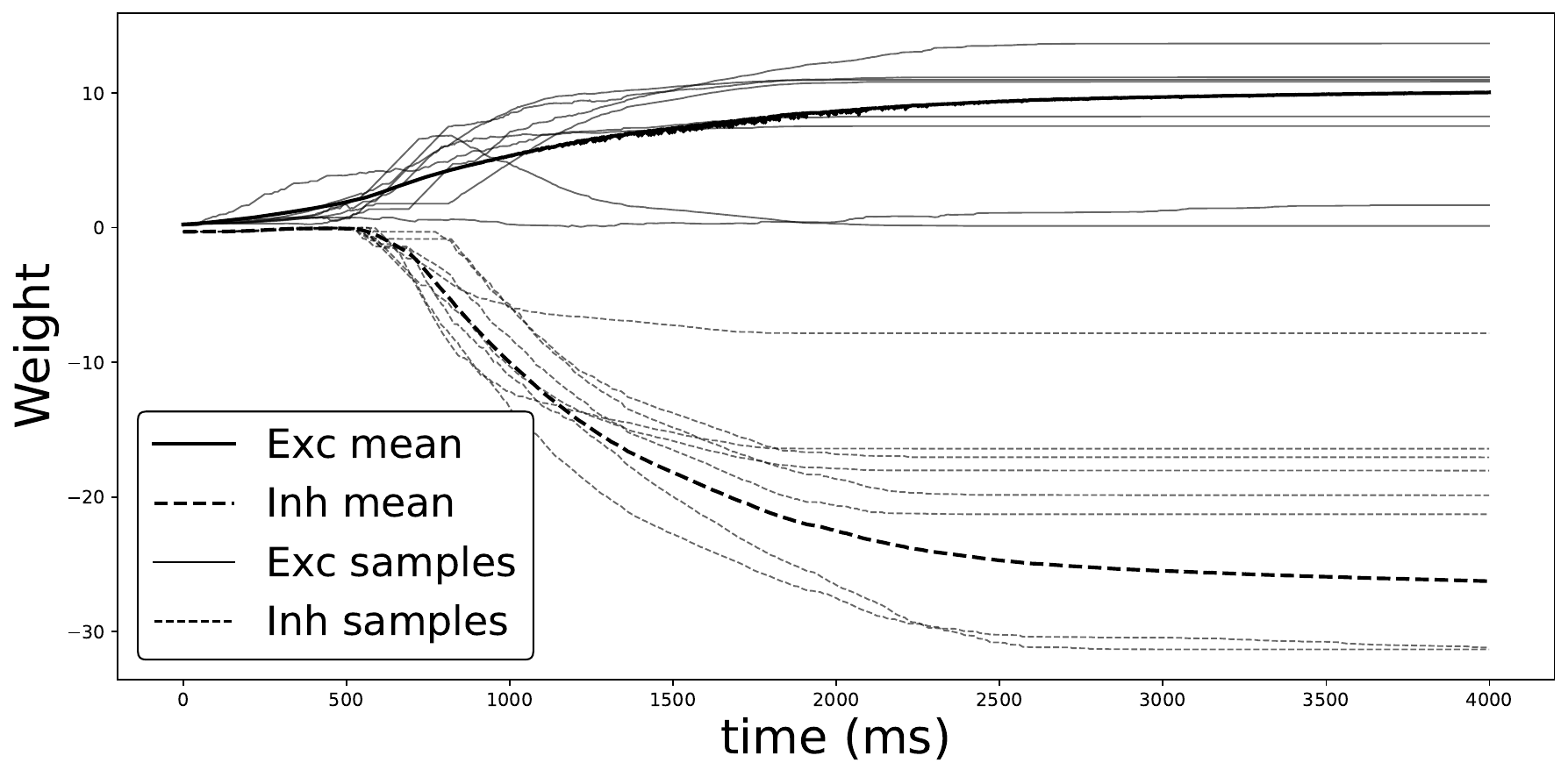}
        \caption{}   % unnumbered caption
        \label{fig:w_epo_no_sleep}
    \end{subfigure}

    \caption{(a) denotes average excitatory and inhibitory weights across batches training with sleep regularization. The grey area capture min and max weights per batch. (b) denotes the same, but without sleep to stabilize learning. (c) shows weight changes across a single batch training session with the sleep protocol active. Along the x-axis, we plot time in milliseconds, and along the y-axis, we plot change in weights. Similar to (d), stippled lines represent inhibitory weights and solid lines excitatory min/max and mean weights.}
    \label{fig:weights_comp}
\end{figure}

We observed that our sleep protocol combined with sign-clamping successfully stabilized the synaptic weights, whereas sign-clamping alone did not (see Figure~\ref{fig:weights_comp}). The sleep protocol achieved high accuracy (mean$=94.93$, min$=82.67$ and max$=100.00$) within the span of 15 batchess, while the \textit{no-sleep} configuration achieved notably worse performance: $(\text{mean}=43.20, \text{min}=24.67 \text{, and max} = 54.67$). In Fig.~\ref{fig:w_evol_sleep}, the weight averages and their min/max spans remain consistent across inhibitory and excitatory groups over training, whilst this is not the case for the \textit{no-sleep} condition (see Fig.~\ref{fig:w_evol_no_sleep}). Within batchess, this trend is also evident: Fig.~\ref{fig:w_epo_sleep} shows the within-batches dynamics: with the sleep protocol active, weights stay within bounded ranges yet still exhibit healthy heterogeneity. Meanwhile, Fig.~\ref{fig:w_epo_no_sleep} demonstrates how weights quickly explode without proper regularization methods in place. 

\begin{figure}[h]
    \centering

    % --- Left figure ---
    \begin{subfigure}[b]{0.45\linewidth}
        \centering
        \includegraphics[width=\linewidth]{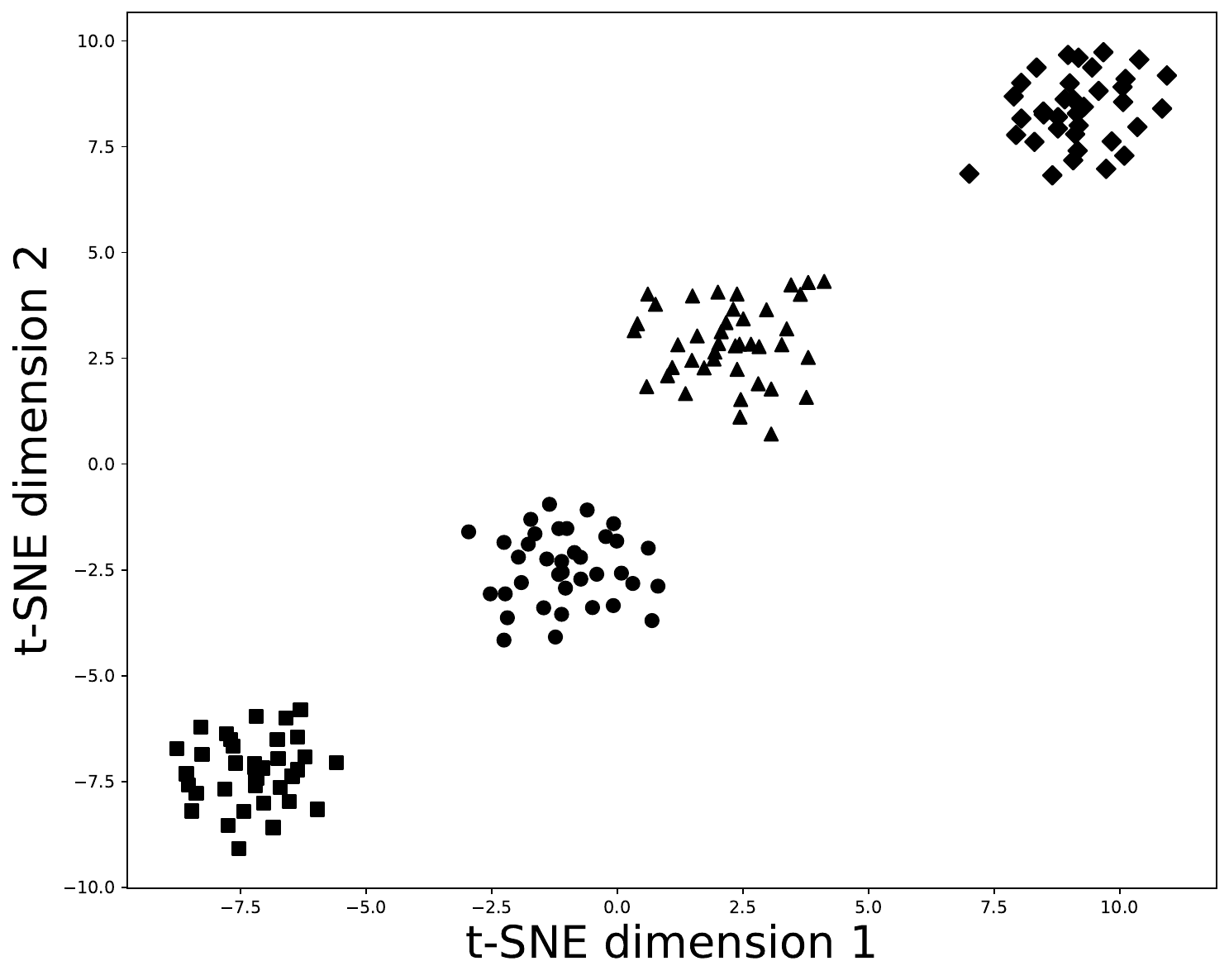}
        \caption{}   % unnumbered caption
        \label{fig:tsne_sleep}
    \end{subfigure}
    \hfill
    % --- Right figure ---
    \begin{subfigure}[b]{0.45\linewidth}
        \centering
        \includegraphics[width=\linewidth]{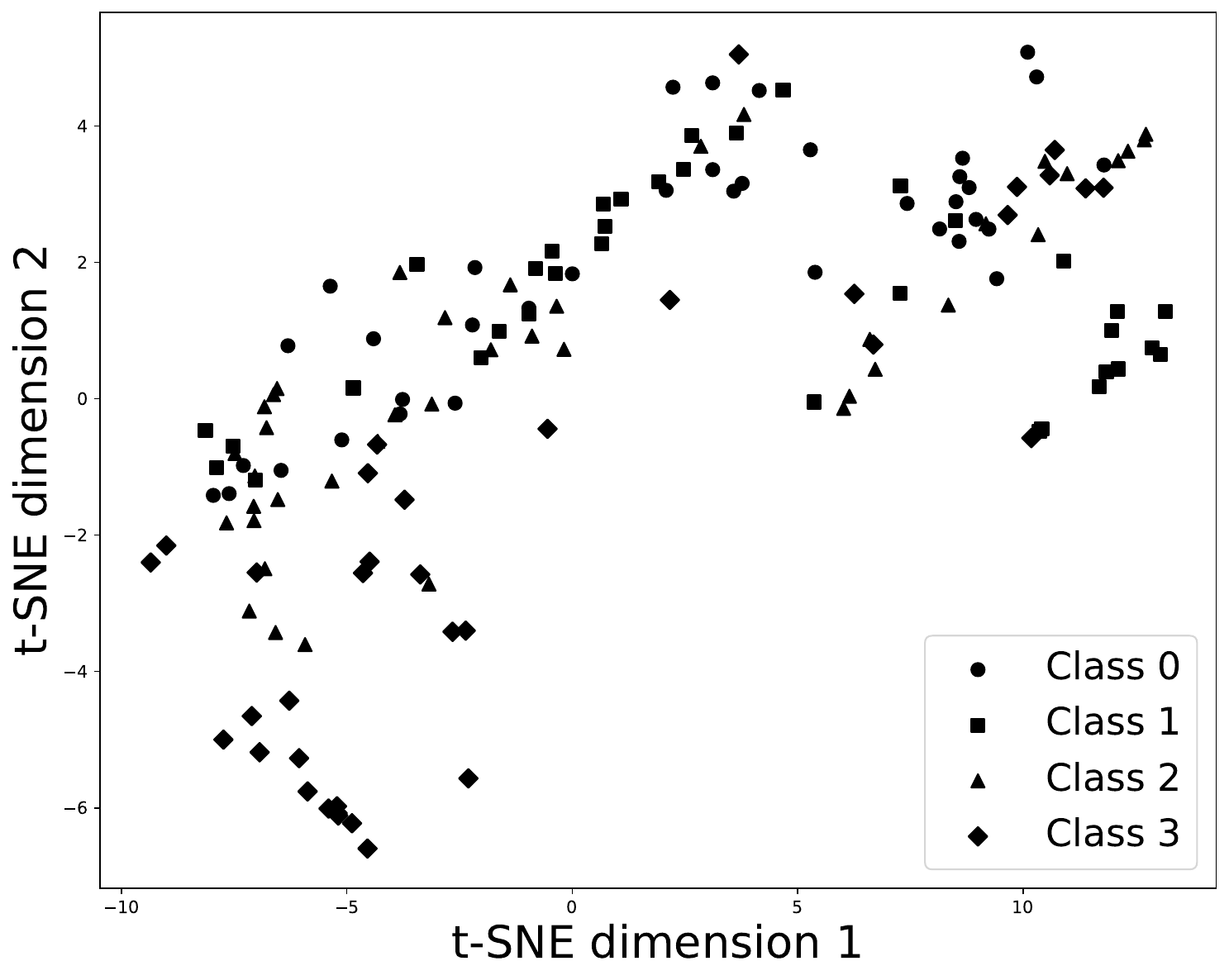}
        \caption{}   % unnumbered caption
        \label{fig:tsne_no_sleep}
    \end{subfigure}

    \caption{Comparison plot of \textit{t}-SNE across sleep (a) and no-sleep (b) training paradigms. Both plots show \textit{t}-SNE visualization of spike-rate vectors binned by training sample. Each geometric shape corresponds to a distinct class, and each point represents a test observation embedded into the \textit{t}-SNE–reduced space. }
    \label{fig:tsne_comparison}
\end{figure}

In contrast, removing the sleep phase leads to rapid instability. Even within a single batch, weights begin to diverge; note the difference in y-axis scales between the sleep and no-sleep conditions, ultimately disrupting learning. This degradation is reflected in Fig.~\ref{fig:tsne_comparison}, where the \textit{no-sleep} model \ref{fig:tsne_no_sleep}, shows impaired clustering structure. Because the sleep protocol restrains weight growth while preserving variability, it yields more stable performance and provides a clear class separation in this example.

\subsection{Surrogate-gradient model}
While the preceding sections focus on stabilizing fully local STDP in a recurrent E/I SNN, it remains unclear whether the observed gains are specific to Hebbian learning or reflect a more general regularization effect. To disentangle these factors, we introduce a second model trained with global error-driven optimization. This surrogate-gradient model provides a strong point of reference for (i) predictive performance under BPTT and (ii) whether the same sleep intervention transfers to gradient-based SNN training.

\subsubsection{Architecture}
The network is a two-layer multilayer perceptron (MLP) with spiking dynamics:

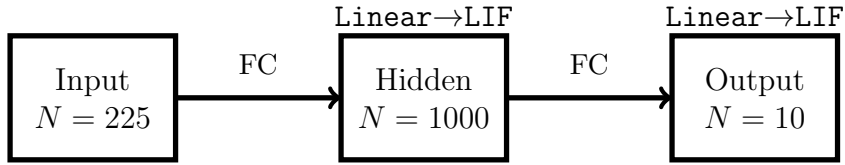
\begin{figure}[h]
\centering
\begin{tikzpicture}[scale=1.1, line width=2pt, font=\normalsize]

% Input box
\draw (0, 0) rectangle (2, 1.5)
node[midway, align=center] {Input\\ $N = 225$};
% Optional: if you truly use Poisson here
%\node[anchor=center] at (1.5, -0.4) {\scriptsize (Poisson encoding)};

% Hidden box
\draw (4, 0) rectangle (6, 1.5) node[midway, align=center] (hidden) 
{Hidden\\ $N = 1000$};
\node[anchor=south, yshift=6pt] at (hidden.north)
{\texttt{Linear}$\rightarrow$\texttt{LIF}};

% Output box
\draw (8, 0) rectangle (10, 1.5) node[midway, align=center] (output)
{Output\\ $N = 10$};

\node[anchor=south, yshift=6pt] at (output.north)
{\texttt{Linear}$\rightarrow$\texttt{LIF}};

% Loss node
%\node[draw, rectangle, minimum width=2.0cm, minimum height=1.1cm, align=center]
%    (loss) at (15.5,0.75) {$\mathcal{L}$\\ \scriptsize CE};

% Arrow from output to loss (forward evaluation)
%\draw[->, line width=2pt] (13,0.75) -- (loss.west);

% Backprop / gradient flow (dashed to avoid looking like recurrence)
%\draw[->, dashed, line width=2pt]
%    (loss.south) .. controls (15.5,-0.8) and (6.5,-0.8) .. (hidden.south)
%    node[midway, below, yshift=-0.2cm] {\small $\nabla W$ (SG-BPTT)};

% Forward arrows
\draw[->, line width=2pt] (2, 0.75) -- (4, 0.75) node[midway, above, yshift=0.15cm] {\small FC};
\draw[->, line width=2pt] (6, 0.75) -- (8, 0.75) node[midway, above, yshift=0.15cm] {\small FC};

\end{tikzpicture}
\caption{Surrogate-gradient baseline: feedforward spiking MLP (225--1000--10) implemented as fully-connected synapses followed by LIF dynamics (\texttt{nn.Linear}$\rightarrow$\texttt{snn.Leaky}).}
\label{fig:architecture_sg}
\end{figure}

implemented as two fully-connected layers (\texttt{Linear}) followed by leaky integrate-and-fire (LIF) neurons (\texttt{snn.Leaky}). The architecture is purely feedforward (i.e. no recurrent connections as in the STDP-model). Each static input image is presented for $T=100$ discrete simulation steps; the temporal dimension arises from the LIF state dynamics rather than time-varying spike-based input.

\subsubsection{Neuron model}
Each spiking layer uses a discrete-time LIF neuron with membrane decay parameter $\beta=0.95$. Let $U(t)$ denote the membrane potential and $S(t)\in\{0,1\}$ the emitted spike at time step $t$. In each layer, the membrane is updated according to the standard LIF recurrence (cf. Eq.~\ref{eq:lif}), and a spike is generated by threshold crossing:
\[
S(t) = H(U(t)-U_{\mathrm{thr}}),
\]
where $H(\cdot)$ is the Heaviside step function.
\subsubsection{Learning rule}
For comparison, we also implement surrogate-gradient (SG) training, in which the non-differentiable spike function is replaced by a smooth proxy during backpropagation. Forward pass: neurons generate spikes via the deterministic LIF threshold crossing rule (Equation~\ref{eq:lif} with a shifted Heaviside step function):
\begin{equation}
S_i(t) = \begin{cases}
    1 & \text{if } U_i(t) \geq U_{thr} \\
    0 & \text{if } U_i(t) < U_{thr}.
\end{cases}
\end{equation}
During backpropagation, the spike nonlinearity is replaced by an arctangent-shaped surrogate $\tilde{S}_i(t)$, providing a smooth approximation to the threshold function and enabling gradient flow. 

\begin{align}
\tilde{S}_i(t) = \frac{1}{\pi}\text{arctan}\left(\pi U_i(t)\frac{\alpha}{2}\right)%, \\
%\frac{\partial \tilde{S}}{\partial U_i(t)} = \frac{1}{\pi}\frac{1}{\left(1+\left(\pi U_i(t)\frac{\alpha}{2}\right)^2\right)}, 
\end{align}
based on \cite{fang_incorporating_2021}, where $\alpha$ is the constant factor which determines the shape of the surrogate gradient. 
Weights are then updated via standard gradient descent on the task loss  ($\mathcal{L}$):
\begin{equation}
w_{ij}(t+1) \gets w_{ij}(t) - \alpha_{\text{SG}} \frac{\partial \mathcal{L}}{\partial w_{ij}(t)},
\end{equation}
where \(\alpha_{\text{SG}}\) is the learning rate for SG training.

\subsubsection{Sleep protocol}
To investigate whether sleep-based regularization can also benefit SG-trained SNNs, we apply the same sleep-wake protocol described above. During wake phases, the network processes task-driven input, and STDP is disabled (gradients govern all updates). During sleep phases, we suppress external input ($I^{\text{task}}=0$), maintain intrinsic noise in the membrane potential, and apply the same stochastic weight decay as in \eqref{eq:decay}.

Unlike Hebbian training, where STDP continues during sleep and enables replay-based consolidation, in SG training, the sleep mechanism provides stochastic perturbations that may help escape suboptimal local minima in the error landscape. However, the multiplicative decay also disrupts the gradient-based weight updates. The net effect, thus, is expected to depend critically on sleep magnitude: at low sleep rates, noise-driven exploration may provide modest benefits by allowing to escape local extrema of the loss; at high rates, decay will likely dominate and degrade completely SG convergence. 

\subsubsection{Evaluation}
Training minimizes a time-accumulated classification objective. For each input sample, the network produces an output spike vector $\mathbf{s}^{(2)}(t)\in\{0,1\}^{C}$ at each simulation step $t\in\{1,\dots,T\}$, where $C=10$ classes. We compute cross-entropy at every step and sum over the stimulus window:
\begin{equation}
\mathcal{L} = \sum_{t=1}^{T} \mathrm{CE}\!\left(\mathbf{s}^{(2)}(t),\, y\right),
\end{equation}
and update the synaptic parameters of the linear layers using Adam.

To obtain a single class prediction from the spiking output, we apply spike-count decoding at the final layer. Let $\mathbf{c}=\sum_{t=1}^{T}\mathbf{s}^{(2)}(t)$ denote class-wise spike counts; the predicted label is
\begin{equation}
\hat{y}=\arg\max_{k\in\{1,\dots,C\}} c_k.
\end{equation}
Accuracy is reported as the fraction of samples with $\hat{y}=y$ on the validation and test sets.

\subsection{MNIST-family experiment}
To formally evaluate our models, we trained them on the MNIST-family of datasets: KMNIST, MNIST, FMNIST and NotMNIST.
\subsubsection{Data}Each dataset contains $60, 000$ images for training and $10, 000$ for testing. Due to computational constraints with our STDP-SNN model, each image is $15\times15$ and each run consisted of a single epoch-run with $6, 000$ randomly drawn images until an even class distribution was achieved for both models. Each batch was subsequently validated with 100 samples and, finally, tested on 1,000 test samples similar to the toy experiment. To compare the effect of sleep, we trained our model on 11 varying sleep rates (0\% to 100\%, with 10\% increments). In addition, to improve statistical confidence while ensuring comparability across sleep rates, we trained the model five times per sleep rate using a fixed set of five random seeds, reused across all sleep conditions. The total number of runs can be expressed as
\begin{equation}
    \text{total runs = seeds $\times$ sleep rates $\times$datasets }
\end{equation}

\FloatBarrier

\begin{figure}[h!]
    \centering
    \includegraphics[width=10cm, height=6cm, trim={0 10 0 0},  clip
      ]{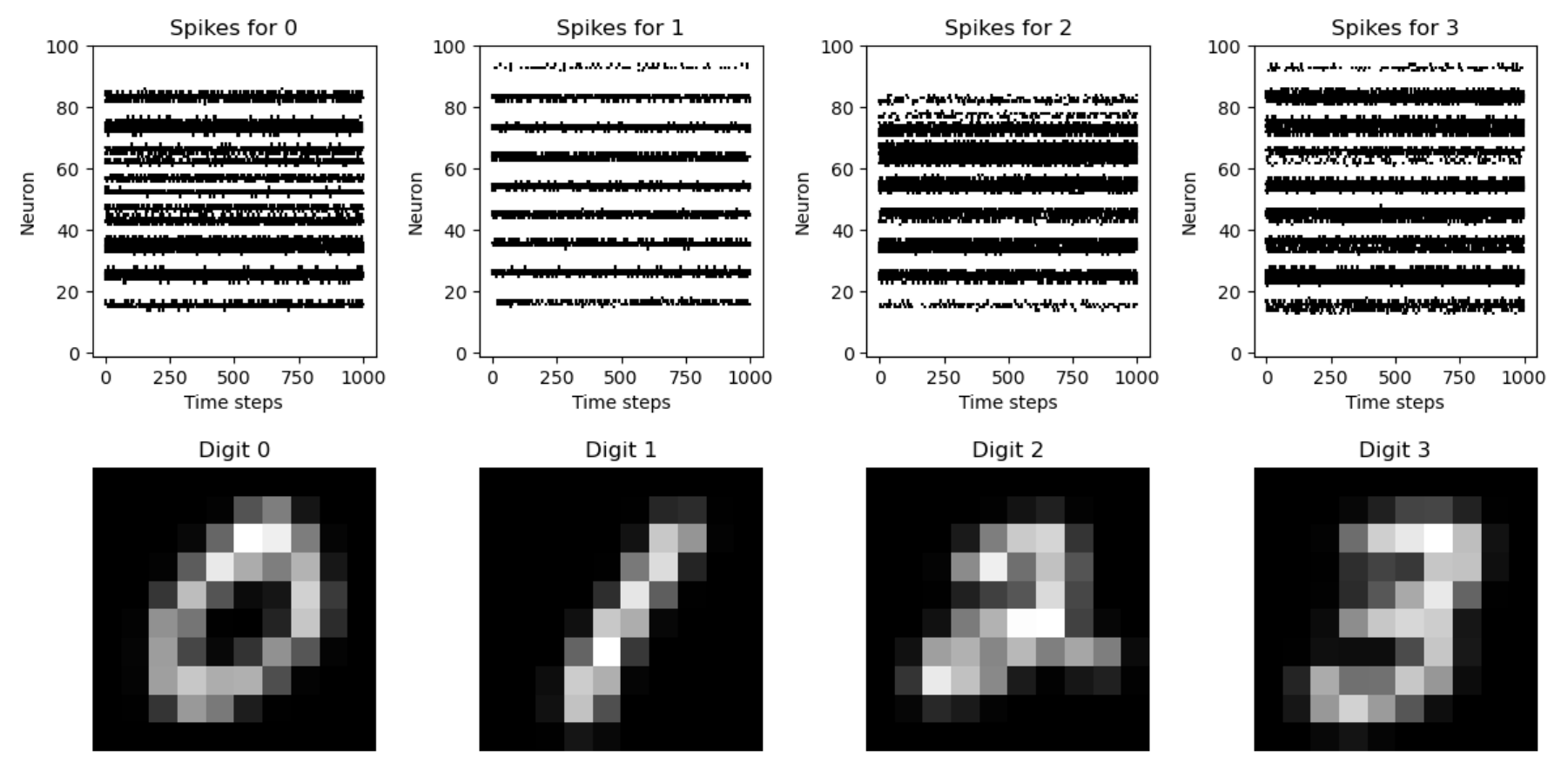}
    \caption{Figure illustrates four examples of training data from the MNIST dataset. The top row demonstrates the resulting spike-encoded training data leveraging snntorch's \textit{spikegen.rate()} function. The neurons are plotted against time, and each dot signifies a spike.}
    \label{figures::spikes_img_comp}
\end{figure}

\subsubsection{GLM-model} To quantify the effect of sleep-based regularization and provide 95\% confidence intervals for the performance, we fit a hierarchical \emph{beta regression} with 
\begin{align}
y_{ijkl} &\sim \mathrm{Beta}(\mu_{ijkl},\phi), \\
\operatorname{logit}(\mu_{ijkl})
&= \beta_0 + \beta_j^{\text{(sleep)}} + \beta_k^{\text{(model)}} + \beta_{jk}^{\text{(sleep$\times$model)}} \notag\\
&\quad + b_l^{\text{(dataset)}} + b_i^{\text{(seed)}} + b_{li}^{\text{(dataset$\times$seed)}} .
\end{align}

where $y_{ijkl}$ is the test accuracy with sleep rate $j$ in model $k$, dataset $l$ and seed $i$.  
Each $\beta_{j}$ quantifies the fixed effect of using sleep rate $j'\text{s}\%$ relative to $0\%$ sleep; all sleep-levels appear as factor variables for direct interpretability.

\section{Results}
The fixed-effect beta regression estimates (Table~\ref{tab:betareg}) statistically substantiate the observed trends. The intercept represents logit accuracy at $0\%$ sleep for STDP-SNN model, while coefficients $\beta_{j=10}$–$\beta_{j=100}$ capture logit accuracy changes at each sleep level relative to this baseline. The strongest effects occur at $\beta_{j=10}$ ($1.56$, $p<.001$) and $\beta_{j=20}$ ($1.59$, $p<.001$), with progressively smaller—though still significant—effects at higher sleep levels. This pattern indicates diminishing returns beyond moderate sleep, consistent with the bias–variance tradeoff observed in other regularization schemes such as $\ell_1$ and $\ell_2$ regularization \citep{tarres_online_2013}.

\begin{figure}[h!]
    \centering
    \includegraphics[width=1\linewidth]{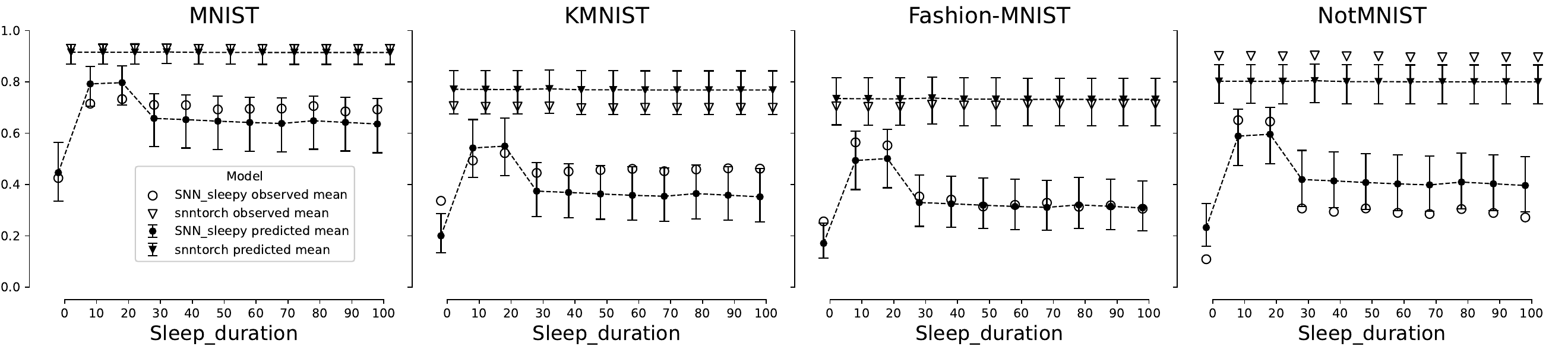}
    \caption{Mean predicted accuracy ($\pm$95\% CI) versus sleep duration for the STDP-SNN model and the SG-SNN model-SG across the MNIST-family benchmarks. Each point shows mean \texttt{glmmTMB} estimate over 5 seeds. The hollow points represent observed accuracies.}
    \label{fig:sleep_benchmark}
\end{figure}

In contrast, the SG-SNN model exhibits qualitatively different behavior. The negative and significant interaction terms closely mirror the positive main effects of sleep, yielding a near-zero net effect across all sleep levels. As a result, the model performance is largely insensitive to sleep duration.

Figure~\ref{fig:sleep_benchmark} reflects this divergence: the SG-SNN model maintain stable, near-optimal accuracy with an optimum at $0\%$ sleep, whereas the STDP-SNN model displays a non-monotonic response, peaking at moderate sleep ($10\%$–$20\%$) and declining at higher rates. This pattern is consistent across all four datasets. Despite the gains from sleep regularization, the SG-STDP model retains higher absolute accuracy overall.

Variance estimates in the lower portion of Table~\ref{tab:betareg} indicate dataset-level heterogeneity, while confirming the robustness of the regularization effect.

{
\setlength{\tabcolsep}{15pt} % Only affects this table
\begin{longtable}{lrrrrl}
    \caption{Mixed effect model with beta regression \label{tab:betareg}}\\
    \toprule
    Effect  & Estimate & Std. Error & \textit{z} value & \textit{p}-value & \\
    \endfirsthead

    \toprule
    Effect  & Estimate & Std. Error & \textit{z} value & \textit{p}-value & \\
    \endhead
        Intercept & -1.098 & .295 & -3.715 & .192 \\[6pt]
        \multicolumn{6}{l}{\textit{Main effect of sleep duration}}\\
        $\beta_{j=10}$ & 1.563 & .156 & 10.026 & $<.001$ & ***\\
        $\beta_{j=20}$ & 1.592 & .156 & 10.212 & $<.001$ & ***\\
        $\beta_{j=30}$ & .875 & .152 & 5.764 & $<.001$ & ***\\
        $\beta_{j=40}$ & .853 & .152 & 5.619 & $<.001$ & ***\\
        $\beta_{j=50}$ & .827 & .152 & 5.444 & $<.001$ & ***\\
        $\beta_{j=60}$ & .804 & .152 & 5.295 & $<.001$ & ***\\
        $\beta_{j=70}$ & .787 & .152 & 5.190 & $<.001$ & ***\\
        $\beta_{j=80}$ & .833 & .152 & 5.497 & $<.001$ & ***\\
        $\beta_{j=90}$ & .806 & .152 & 5.298 & $<.001$ & ***\\
        $\beta_{j=100}$ & .778 & .152 & 5.118 & $<.001$ & ***\\[6pt]
        \multicolumn{6}{>{\arraybackslash}p{\linewidth}}{\textit{Main effect of model (reference = the STDP-SNN model)}}\\
        Model: SG-SNN & 2.614 & .167 & 15.657 & $<.001$ & ***\\[6pt]
        $\beta^{\text{(sleep×model)}}_{j=10}$ & -1.566 & .230 & -6.797 & $<.001$ & ***\\
        $\beta^{\text{(sleep×model)}}_{j=20}$ & -1.599 & .230 & -6.939 & $<.001$ & ***\\
        $\beta^{\text{(sleep×model)}}_{j=30}$ & -.863 & .228 & -3.789 & $<.001$ & ***\\
        $\beta^{\text{(sleep×model)}}_{j=40}$ & -.864 & .228 & -3.797 & $<.001$ & ***\\
        $\beta^{\text{(sleep×model)}}_{j=50}$ & -.838 & .228 & -3.682 & $<.001$ & ***\\
        $\beta^{\text{(sleep×model)}}_{j=60}$ & -.819 & .228 & -3.600 & $<.001$ & ***\\
        $\beta^{\text{(sleep×model)}}_{j=70}$ & -.802 & .228 & -3.528 & $<.001$ & ***\\
        $\beta^{\text{(sleep×model)}}_{j=80}$ & -.849 & .227 & -3.731 & $<.001$ & ***\\
        $\beta^{\text{(sleep×model)}}_{j=90}$ & -.821 & .228 & -3.605 & $<.001$ & ***\\
        $\beta^{\text{(sleep×model)}}_{j=100}$ & -.793 & .228 & -3.483 & $<.001$ & ***\\
        \midrule
        \multicolumn{2}{l}{Random effect variance}\\
        Groups & Name & Variance & Std.Dev.& &\\
        Dataset & (Intercept) & .289 & .538 & &\\
        Seed & (Intercept) & $< .001$ & $< .001$& &\\
        Dataset:Seed & (Intercept) & .037 & .192& &\\
        
        \bottomrule
    %\caption*{Significance levels: * $p<.05$, ** $p<.01$, *** $p<.001$.}
\end{longtable}
}
\FloatBarrier

\section{Discussion}

This study shows that homeostatic, sleep-like regularization can substantially improve learning stability and generalization in locally trained SNNs—but only when applied within a narrow and well-defined regime. Across the full MNIST family of benchmarks, we find consistent evidence that modest sleep durations yield measurable performance gains, while excessive sleep degrades learning. Together, these results present a mechanistically grounded and statistically robust account of how sleep-like processes interact with local synaptic plasticity.

The primary role of interleaved sleep phases is to prevent pathological weight growth and catastrophic interference. In recurrent SNNs trained with STDP, synaptic updates accumulate without an intrinsic normalization mechanism, leading to runaway weight magnitudes even in networks with balanced excitatory and inhibitory populations. Introducing periodic, power-law weight decay during sleep counteracts this instability by constraining synapses to remain within functional ranges. This restores effective plasticity and stabilizes learning dynamics over long training horizons.

Sleep also contributes beyond mere weight control. During sleep phases, externally driven input is removed and replaced by low-amplitude Gaussian noise, inducing spontaneous network activity. This background activity leads neurons to stochastically reactivate previously learned assemblies. Because STDP remains active during these quiescent periods, synaptic structure can be selectively reinforced without interference from new sensory patterns. This implicit replay mechanism provides a form of consolidation that may support generalization.

Crucially, our results show that the benefits of sleep are sharply bounded. For the STDP-SNN model, only mild regularization—approximately $10\%-20\%$—improves performance. Beyond this regime, accuracy decreases monotonically across all datasets. The beta-regression analysis in Table~\ref{tab:betareg} quantifies this effect: while $\beta_{j=10}$ and $\beta_{j=20}$ are strongly positive, coefficients associated with higher sleep levels progressively diminish in magnitude and ultimately become detrimental relative to the optimal window. This pattern mirrors behavior seen in classical regularization methods (e.g., $\ell_1$ or $\ell_2$ penalties), where overly strong regularization increases bias and impairs performance. 

Finally, our findings highlight an important interaction between learning rule and sleep efficacy. In Hebbian SNNs, sleep decay and STDP operate on compatible local activity statistics, jointly providing weight homeostasis and memory stabilization. In contrast, surrogate-gradient models exhibit a fundamentally different response. At very low sleep rates, stochastic perturbations during sleep can occasionally aid exploration of the loss landscape, yielding small performance improvements. However, as sleep duration increases, global renormalization increasingly conflicts with gradient-based optimization, potentially disrupting the coordination between error signals and weight updates. Yet, empirically, these opposing effects largely cancel out, resulting in performance that remains effectively invariant to sleep duration across the MNIST-family benchmarks.

Our implementation is amenable to neuromorphic deployment. All regularization and plasticity operations are fundamentally local, require no additional inference-time cost, and impose only $O(1)$ per-weight computational burden during sleep, thus being  fully compatible with event-driven hardware. 

\subsection{Limitations} Several limitations should be noted. First, the number of simulation steps per sample differs between tasks (1000 ms for the geometric dataset versus 100 ms for the MNIST-family datasets). This reduction reflects computational constraints, as the MNIST-family experiments required multiple evaluations per sleep rate (five runs across eleven sleep levels), rendering longer per-sample simulations prohibitively expensive.

Second, the SG-SNN differs fundamentally from the STDP-SNN in both learning dynamics and data representation. While the STDP-SNN relies exclusively on local spike-timing-dependent plasticity during both wake and sleep, the SG-SNN is trained via surrogate-gradient backpropagation during wake phases, with sleep introducing only stochastic weight decay and auxiliary trace-based updates. Moreover, the SG-SNN processes static pixel-valued inputs presented as continuous currents, whereas the STDP-SNN operates on spike-encoded sensory input. The two models also employ different decoding and accuracy estimation procedures. As a result, observed performance differences cannot be attributed solely to the presence or absence of sleep-based regularization, but must be interpreted in light of the distinct learning signals and representational assumptions underlying each model.

Most importantly, predictive performance of the suggested models is far from state-of-the-art, and our work should be see as foundational for small incremental improvements of Hebbian learning within spiking neural networks, which still struggle to match the performance of regular artificial neural networks. Our work is no exception in this sense. 

\subsection{Future work}Several promising directions emerge from this study. First, embedding the proposed sleep regularization into deeper SNN architectures and evaluating it on larger-scale datasets would help assess scalability, robustness, and practical limitations. Extending these experiments to biologically inspired multimodal data is particularly well aligned with the biologically grounded nature of the proposed approach.

Because the sleep protocol relies on noisy membrane activity to preserve relative synaptic strengths, an important next step is to disentangle how noise contributes to this stabilization. In particular, systematically varying noise levels across different sleep rates could reveal interaction effects and clarify the conditions under which sleep most effectively supports learning.

Beyond noise-driven mechanisms, it would be valuable to examine how sleep interacts with other forms of homeostatic regulation, such as intrinsic plasticity, attentional modulation, or reward-based learning. Studying these interactions may shed light on whether sleep complements or competes with alternative stabilization strategies.

Finally, deploying the proposed models on neuromorphic hardware would provide a compelling test of their real-world applicability, while also enabling direct comparisons with other locally operating, continuous-time learning algorithms. 

\subsection{Conclusion}
We showed that sleep-inspired homeostatic regularization can stabilize and improve locally trained STDP-based spiking neural networks, but only within a narrow operating regime. Across the MNIST-family benchmarks, moderate sleep fractions (10--20\%) consistently improved performance, whereas the same intervention produced no systematic benefit in surrogate-gradient training. Using mechanistic motivation together with systematic sweeps and mixed-effects beta regression, we characterized both the benefits and limits of local sleep-based renormalization.

Although our models do not match state-of-the-art accuracy, the proposed protocol preserves strict locality and has low computational overhead, making it well suited for neuromorphic and on-chip learning. More broadly, the results suggest that sleep-like renormalization is a practical ingredient for controlling pathological weight dynamics in Hebbian SNNs, and a promising direction for biologically grounded, robust learning systems.

\newpage

\appendix
\section{Hyperparameters}

\begin{longtable}{m{3cm}m{2cm}m{4cm}m{2cm}}
\caption{Key tuning parameters for all experiments with \textit{the STDP-SNN model}}
\label{tab:hparams_snn_stdp}\\
\toprule
Component & Parameter & Descriptor & Value \\
\midrule
\endfirsthead

\toprule
Component & Parameter & Descriptor & Value \\
\midrule
\endhead

Network           &&&\\
                  & $N_{\text{in}}$ & Input neurons & 225 \\
                  & $N_{\text{exc}}$ & Excitatory neurons & 200 \\
                  & $N_{\text{inh}}$ & Inhibitory neurons & 50 \\
                  & $P_{\text{in}\to\text{exc}}$ & Wiring prob. & 15\% \\
                  & $P_{\text{exc}\to\text{exc}}$ & Wiring prob. & 10\% \\
                  & $P_{\text{exc}\to\text{inh}}$ & Wiring prob. & 20\% \\
                  & $P_{\text{inh}\to\text{exc}}$ & Wiring prob. & 25\% \\
                  & $W_{\text{in}\to\text{exc}}$  & Synaptic weight & 0.10 \\
                  & $W_{\text{exc}\to\text{exc}}$ & Synaptic weight &  0.15\\
                  & $W_{\text{inh}\to\text{exc}}$ & Synaptic weight & 0.30 \\
                  & $W_{\text{exc}\to\text{inh}}$ & Synaptic weight & $-0.30$ \\
STDP              &&&\\
                  & $\eta_{\text{exc}}$ & STDP rate & $5\times10^{-4}$ \\
                  & $\eta_{\text{inh}}$ & iSTDP rate & $5\times10^{-4}$ \\
                  & $A_+$ & Potentiation & 0.5 \\
                  & $A_-$ & Depression & 0.3 \\
                  & $\tau_+$ & LTP time constant & 10 ms \\
                  & $\tau_-$ & LTD time constant & 7.5 ms \\
Neuro-dynamics   &&&\\
                  & $\Delta t$ & Time constant & 1 ms\\
                  & $\tau_m$ & Membrane time const & 30 ms \\
                  & $U_{\text{rest}}$ & Resting potential & $-70$ mV \\
                  & $U_{\text{reset}}$& Reset potential  & $-80$ mV \\
                  & $U_{max}$ & Max potential & $40$ mV \\
                  & $U_{min}$ & Min potential & $-100$ mV \\
                  & $R_m$ & Membrane resistance & $30$ mV \\
                  & $\lambda_{\mu}$ & Membrane noise mean & $0 $mV\\
                  & $\lambda_{\sigma}$ & Membrane noise var & $3 $mV\\
                  & $\alpha$ & LIF threshold baseline & $-55$ mV \\
                  & $\tau_{\text{th}}$ & Threshold tc     & 100 ms \\
                  & $\delta$ & Thresh. jump                & 3 \\
Sleep/homeostasis &&&\\
                  & $\beta$ & Sleep percentage            & 10\% \\
                  & $\alpha_{\text{base}}$& Sleep lower bound & 1.0 \\
                  & $\lambda$ & Sleep decay power               & 0.9997 \\
                  & $w_{\text{tgt}}$ & Target weight   & $\mid0.2\mid$ \\
Data processing &&&\\
                  & $N_{\text{train}}$ & Training samples & 6000 \\
                  & $N_{\text{test}}$ & Test samples & 1000 \\
                  & $N_{\text{val}}$ & Validation samples & 100 \\
                  & $S_{\text{image}}$ & Image resize dims & $(15, 15)$ \\
                  & $f_{\max}$ & Max spiking rate                 & 1000 Hz \\
                  & $T_{\text{image}}$ & Stimulus duration         & 100 ms \\
                  & $B$ & Batch size & 400 \\
                  & $N_{\text{batches}}$ & Number of batches & 15 \\
Early Stopping &&&\\
                  & $p_{\text{patience}}$ & Training patience & 0.2 \\
\bottomrule
\end{longtable}

\begin{longtable}{m{3cm}m{2cm}m{4cm}m{2cm}}
\caption{Key tuning parameters for all experiments with \textit{the SG-SNN model}}
\label{tab:hparams_the SG-SNN model}\\
\toprule
Component & Parameter & Descriptor & Value \\
\midrule
\endfirsthead

\toprule
Component & Parameter & Descriptor & Value \\
\midrule
\endhead

Network           &&&\\
                  & $N_{\text{in}}$ & Input neurons & 225 \\
                  & $N_{\text{exc}}$ & Hidden neurons & 1000 \\
                  & $N_{\text{exc}}$ & Output neurons & 10 \\
                  & $P_{\text{in}\to\text{hid}}$ & Wiring prob. & 100\% \\
                  & $P_{\text{hid}\to\text{out}}$ & Wiring prob. & 100\% \\
                  & $W_{\text{in}\to\text{hid}}$  & Synaptic weight & 1.0 \\
                  & $W_{\text{hid}\to\text{out}}$ & Synaptic weight &  1.0\\
STDP              &&&\\
                  & $\eta_{\text{exc}}$ & STDP rate & $1\times10^{-3}$ \\
                  & $\eta_{\text{inh}}$ & iSTDP rate & $1\times10^{-3}$ \\
                  & $A_+$ & Potentiation & 0.01 \\
                  & $A_-$ & Depression & 0.012 \\
                  & $\tau_{\text{pre}}$ & Pre-syn trace const. & 20 ms \\
                  & $\tau_{\text{post}}$ & Post-syn trace const. & 20 ms \\
Neuro-dynamics   &&&\\
                  & $\Delta t$ & Time constant & 1 ms\\
                  & $\tau_m$ & Membrane decay & 0.95 ms \\
                  & $U_{\text{reset}}$& Reset potential  & 0 \\
                  & $\lambda_{\mu}$ & Membrane noise mean & $0$ \\
                  & $\lambda_{\sigma}$ & Membrane noise var & $0.5 $\\
                  & $\alpha$ & LIF threshold baseline & $1$ mV \\
Sleep/homeostasis &&&\\
                  & $\beta$ & Sleep percentage            & [0:100\%] \\
                  & $\alpha_{\text{base}}$& Sleep lower bound & 1.0 \\
                  & $\lambda$ & Sleep decay power               & 0.999 \\
                  & $w_{\text{tgt}}$ & Target weight   & $0.2$ \\
Input encoding    &&&\\
                  & $f_{\max}$ & Max spiking rate                 & 1000 Hz \\
                  & $T_{\text{image}}$ & Stimulus duration         & 100 ms \\
Data pre-processing &&&\\
                  & $N_{\text{train}}$ & Training samples & 6000 \\
                  & $N_{\text{test}}$ & Test samples & 1000 \\
                  & $N_{\text{val}}$ & Validation samples & 100 \\
                  & $S_{\text{image}}$ & Image resize dims & $(15, 15)$ \\
                  & $f_{\max}$ & Max spiking rate                 & 1000 Hz \\
                  & $T_{\text{image}}$ & Stimulus duration         & 100 ms \\
                  & $B$ & Batch size & 400 \\
                  & $N_{\text{batches}}$ & Number of batches & 15 \\
Optimization &&&\\
                  & $\eta_{\text{Adam}}$ & Adam learning rate & $5\times10^{-4}$ \\
                  & $\beta_1$ & Adam first decay rate & 0.9 \\
                  & $\beta_2$ & Adam sec. decay rate & 0.999 \\
Early Stopping &&&\\
                  & $p_{\text{patience}}$ & Patience fraction & 0.2 \\
                  & $\Delta_{\text{min}}$ & Min. improv. thresh. & 0.001 \\
                  
\bottomrule
\end{longtable}

\bibliographystyle{plainnat}
\bibliography{refs}
\end{document}